\documentclass[examplefnt,chapter]{nowfnt}

\usepackage[utf8]{inputenc}
\usepackage{tikz}
\usepackage[framemethod=TikZ]{mdframed}
\usepackage{svg}
\usepackage{subfig}
\usepackage{enumitem,multicol}
\usepackage{multirow}
\usepackage{array}
\usepackage{booktabs}
\usepackage{colortbl}
\usetikzlibrary{trees,shapes,arrows,positioning}
\usepackage{graphicx}
\usepackage{lipsum}
\usepackage{amsmath}
\usepackage{amsthm}
\usepackage{amsfonts}
\usepackage{cleveref}
\usepackage{mathtools}
\usepackage{tabularx}

\AtBeginDocument{\mainmatter}

\begin{document}

\newcommand{\anton}[1]{{\color{red}[Anton: #1]}}

\chapter{Large Language Model Driven Recommendation}
\label{ch:nlp}

\vspace*{-23em}  
{\noindent\small 
\linespread{0.8}\selectfont
This is a preprint of Chapter 4 in the upcoming book \textit{Recommendation with Generative Models}
by Yashar Deldjoo, Zhankui He, Julian McAuley, Anton Korikov, Scott Sanner, Arnau Ramisa, Rene Vidal, Mahesh Sathiamoorthy, Atoosa Kasrizadeh, Silvia Milano, and Francesco Ricci.}

\vspace{10em}

{\noindent\small
\textbf{Chapter Authors:} Anton Korikov, Scott Sanner}

\vspace{5em}


\begin{abstract}
While previous chapters focused on recommendation systems (RSs) based on standardized, non-verbal user feedback such as purchases, views, and clicks -- the advent of LLMs has unlocked the use of natural language (NL) interactions for recommendation. This chapter discusses how LLMs' abilities for general NL reasoning present novel opportunities to build highly personalized RSs -- which can effectively connect nuanced and diverse user preferences to items, potentially via interactive dialogues. To begin this discussion, we first present a taxonomy of the key data sources for language-driven recommendation, covering item descriptions, user-system interactions, and user profiles. We then proceed to fundamental techniques for LLM recommendation, reviewing the use of encoder-only and autoregressive LLM recommendation in both tuned and untuned settings. Afterwards, we move to multi-module recommendation architectures in which LLMs interact with components such as retrievers and RSs in multi-stage pipelines. This brings us to architectures for conversational recommender systems (CRSs), in which LLMs facilitate multi-turn dialogues where each turn presents an opportunity not only to make recommendations, but also to engage with the user in interactive preference elicitation, critiquing, and question-answering.         
\end{abstract}

\section{Introduction}
The advent of LLMs has enabled conversational, natural language (NL) interactions with users -- while also unlocking the rich NL data sources within recommendation systems (RSs) such as item descriptions, reviews, and queries. These advances create the opportunity for highly personalized RSs which harness the general reasoning abilities of LLMs to accommodate diverse and nuanced user preferences through customized recommendations and interactions. 
Such NL-based personalization contrasts starkly to the ID-based RSs described in Chapters 2 and 3 which are highly specialized for standard recommendation tasks (e.g., rating prediction, sequential recommendation) and require large volumes of non-textual interaction data -- though many synergies between these two paradigms are possible, as discussed below.

\subsection{Natural Language vs. Non-Textual Interaction Data}
\begin{figure}[t]
    \centering
    \includegraphics[width = \linewidth]{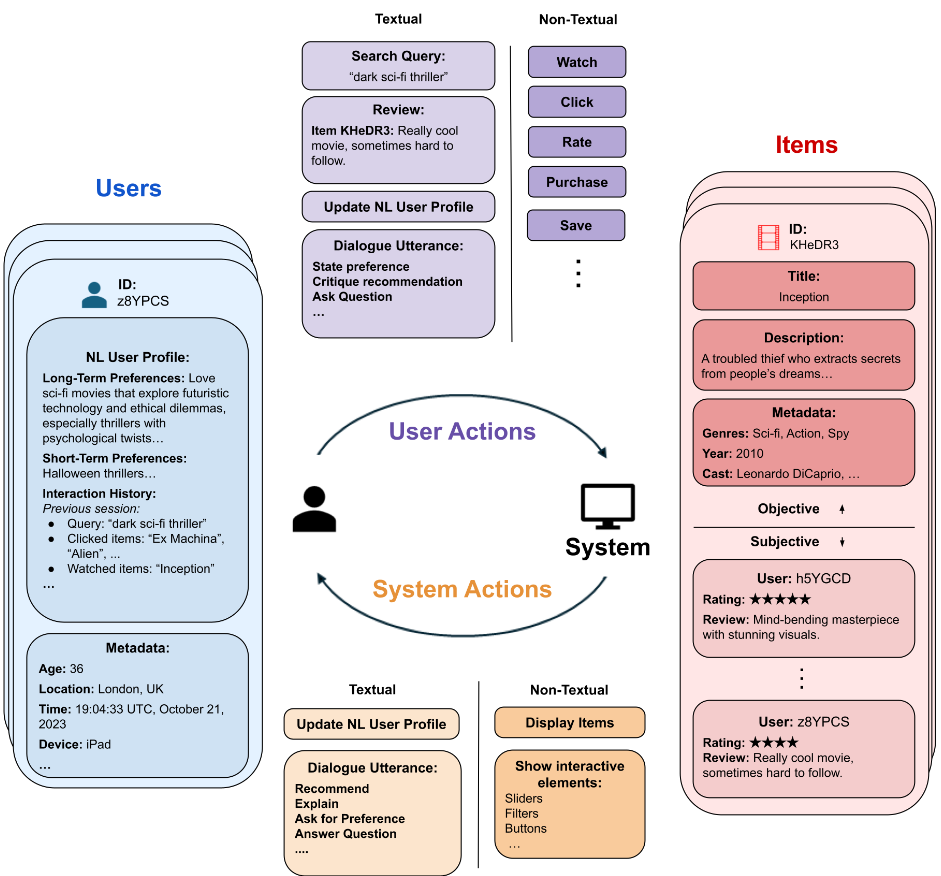}
    \caption{Sources of data in LLM-driven RSs, including item descriptions (right), user-system interactions (top/bottom), and user profiles (left).}
    \label{fig:ch4_data}
\end{figure}

Both textual and non-textual data are important in this chapter -- Figure \ref{fig:ch4_data} illustrates how such data can represent key RS information, covering items, users, and system interactions. This figure is discussed in detail in Section \ref{sec: NL in RS}. 

\subsubsection{Non-textual Interaction Data:} 
On one hand, non-textual user-item interactions such as purchases, views, clicks, or ratings facilitate the collection of large volumes of data in fixed formats (e.g., rating matrices, item ID sequences) and enable the large-scale training of conventional 
RSs -- namely, collaborative filtering (CF) and content-based filtering (CBF) systems (c.f. Ch 2). On the other hand, these non-textual data forms only provide a 
narrow, highly standardized view of user-system interactions -- limiting the degree of personalization that can be achieved.

\subsubsection{Natural Language in Recommendation:} 
In contrast, while NL is a much more complex medium, it is also far more expressive and constitutes a unified format to represent nuanced information about items, preferences, and user-system interactions \citep{geng2022recommendation}. Many RSs already contain an abundance of NL data in the form of item descriptions, reviews, and user query histories, and systems can also generate new NL representations from interaction histories by using LLMs or templates \citep{radlinski2022natural,sanner2023large, zhou2024language}. NL conversational recommendation dialogues are also emerging as a key source of data, capturing a variety of user and system intents in multi-turn interactions (c.f. Sec. \ref{sec:ConvRec}). 



\subsection{General vs. Specialized Recommendation Reasoning}

\subsubsection{Task-specific Reasoning with Conventional RSs}

Conventional
RSs must be trained on large amounts of task-specific interaction data  -- making them highly-specialized tools that are typically optimized for either rating prediction or top-$k$, sequential, or page-wise recommendation. While their performance on \textit{offline} benchmarks for these tasks is generally very strong (c.f. Ch. 6), these specialized systems are limited by the inability of standardized interaction data (e.g., purchases, views, clicks, etc.) to capture nuanced and diverse user preferences. 
These systems also often require considerable data engineering efforts and design time to deploy. 

However, despite these limitations, conventional RS remain powerful and scalable methods for predicting future interactions, able to work with millions of users and items. Therefore, as will be discussed in Sections \ref{sec:RARec}-\ref{sec:ConvRec}, many researchers are actively studying the integration of conventional RSs as specialized sub-components 
within larger, LLM-powered system architectures (e.g., \cite{friedman2023leveraging,hou2023large,zhang2023collm}).


\subsubsection{General Recommendation Reasoning with LLMs}
In contrast to conventional RSs, the pretraining of LLMs on large text corpora provides them with emergent abilities for \textit{general} reasoning -- with LLMs achieving impressive performance on many diverse and previously unseen tasks
\citep{sparks_of_agi}. Through pretraining, LLMs have internalized fine-grained knowledge about a wide range of entities, human preferences, and interaction patterns -- knowledge which could enhance RS personalization while reducing data and design time requirements.   

\begin{figure}[h!]
    \centering
    \includegraphics[width = \textwidth]{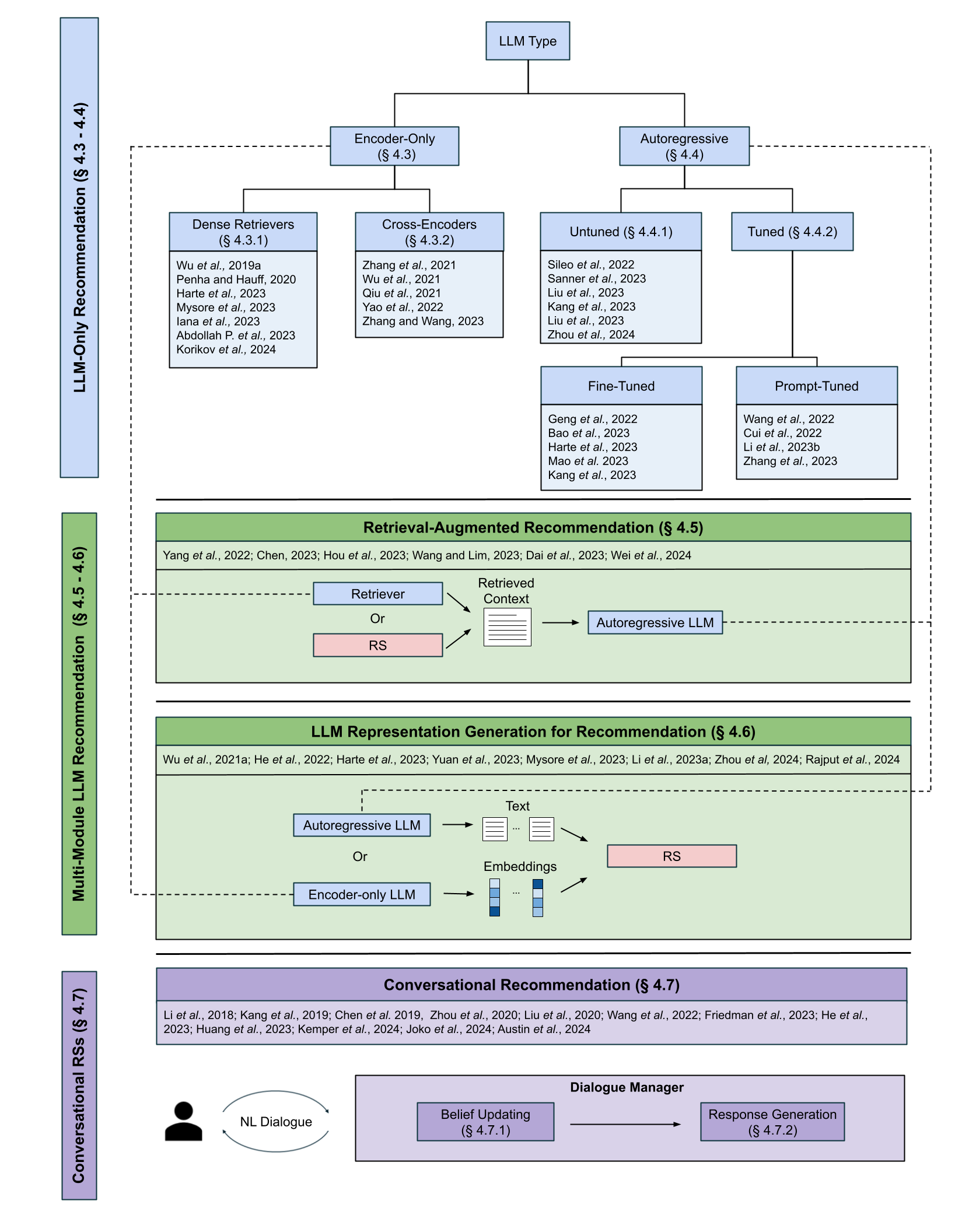}
    \caption{Outline of key techniques and publications in LLM-driven RSs.}
    \label{fig:ch4_outline}
\end{figure}

\paragraph{Recommendation and Explanation} Firstly, 
LLMs can use NL data (e.g., descriptions of items and a user's preferences) to make recommendations -- either by generating text (c.f. Sec \ref{sec:LLM gen rec}) or via embedding-based item scoring (c.f. Sec \ref{sec:encoder-only LLM}). Further, LLMs can generate textual explanations (c.f. \ref{sec:gen expl}) to help users understand why recommendations were made and enable better user feedback, such as NL critiques or follow-up questions.  

\paragraph{Conversational Recommendation} LLMs can also drive NL recommendation dialogues where users \textit{interactively} convey various intents, including: stating and refining preferences, critiquing recommendations, asking questions, or engaging in trade-off negotiations (c.f. Sec \ref{sec:ConvRec}). Conversational recommendation systems (CRSs) can facilitate such complex dialogues by leveraging LLMs
to generate a variety of personalized responses, including recommendations and explanations, answers to questions, and requests for more information. Thus, the general reasoning abilities of LLMs provide opportunities to better personalize not only recommendations, but also user-system interaction sessions more broadly. 

\paragraph{Limitations of LLM-Driven Recommendation} Unfortunately, these LLM-driven opportunities for RSs also come with new risks, biases, and limitations, as discussed further in Chapter 7. Firstly, LLMs may hallucinate, generating outputs which are incorrect or misleading \citep{ji2023survey} which creates significant risks in settings where reliability is key. More broadly, our ability to control LLM behaviour is limited: while prompt engineering and fine-tuning influence outputs, neither approach achieves total control \citep{mialon2023augmented}. 
More optimistically however, this chapter also discusses approaches to mitigate some of these limitations, including through retrieval-augmented generation (RAG) and external tool calls to improve system control and reliability (c.f. Sec \ref{sec:RARec}-\ref{sec:ConvRec}).

\subsection{Chapter Outline}
Before diving into recommendation methodologies, we first present a structured overview of NL data sources for describing items, users, and interactions in Section \ref{sec: NL in RS}. Then, as summarized in Figure \ref{fig:ch4_outline}, the subsequent sections cover key techniques and research in LLM-driven recommendation. First, we describe single turn LLM recommendation, covering the use of both encoder-only and autoregressive models (c.f. Sec. \ref{sec:encoder-only LLM}-\ref{sec:LLM gen rec}). The next two sections focus on synergies between LLMs, conventional RSs, and information retrieval in a discussion of RAG (c.f. Sec. \ref{sec:RARec}) and LLM-based representation generation (c.f. Sec. \ref{sec: LLM 4 RS Inputs}). Finally, we look at architectures for conversational recommendation, surveying various approaches for managing multi-turn and multi-intent dialogues (c.f. Sec \ref{sec:ConvRec}).  

\section{Data Sources in LLM-Driven RSs} \label{sec: NL in RS}
The use of language in recommendation is not new. For instance, text has long been leveraged for content-based recommendation \citep{lops2011content}, key-phrase explanations \citep{mcauley2013hidden, wu2019deep}, and metadata-driven Dialogue State Tracking (DST, \cite{yan2017building}). 
However, LLM's have enabled far more advanced NL reasoning about items, users, and their interactions \citep{geng2022recommendation}, creating opportunities for more nuanced and interactive RS personalization. This section thus outlines the primary data sources which could be used by LLM-era RS, covering item data, interaction data, and user profiles, summarized in Figure \ref{fig:ch4_data}. 

\subsection{Item Text} \label{sec:item text}
The right side of Figure \ref{fig:ch4_data} illustrates data sources for textual item representations, including titles, descriptions, metadata, and reviews.

\subsubsection{Titles} In most settings, items are assigned a \textit{title}: a short, descriptive text span which aims to distinguish the item. A title's capacity to represent an item can vary greatly with domain and item popularity: for example while some hit movies and books can be summarized in one or two words (e.g., \textit{``Titanic (1999)''}, \textit{``Dune (1965)''}), less unique and less popular items (e.g., articles of clothing) do not have such information-dense titles. Items may even entirely lack titles in certain domains, such as user-generated social media video recommendation, though in these cases, it may be possible to generate titles from visual content and metadata with a multimodal LLM (c.f. Ch. 5).

\subsubsection{Descriptions} Items may also be associated with longer and more detailed NL content which we call \textit{descriptions}, which may be human-written or LLM generated (e.g. \cite{acharya2023llm,wei2024llmrec,li2023personalized}). The length and format of descriptions can vary greatly across domains, ranging from non-existent (e.g., social media videos), to short summaries (e.g., eShopping products), to many pages long (e.g., news articles and books).


\subsubsection{Metadata} Item metadata often contains \textit{structured} information about item attributes such as product categories, brands, technical specifications, price, release date, and so on. It can contain various types of information including numerical, categorical, temporal, geographical, and visual data.

\subsubsection{Reviews} A plentiful source of user-generated text are often reviews -- which express nuanced opinions across a diverse range of item attributes and user experiences. However, reviews are often highly subjective, especially when describing ``soft'' attributes such as ``inexpensive'', ``funny'', or ``safe'' 
\citep{radlinski2022subjective,balog2021interpretation}.

\subsection{Interaction Data} \label{sec:textual interactions}
The top and bottom of Figure \ref{fig:ch4_data} illustrate sources of user-system interaction data, including both non-textual interactions (e.g., clicks, purchases, etc.) and NL interactions such as queries, dialogue utterances, and reviews (discussed in the previous section).

\subsubsection{Verbalizing Non-Textual Interactions}
Conventional non-textual user-item interaction history -- such as views, clicks, likes, purchases, and ratings -- can easily be represented as text. For this, a simple template may be sufficient -- for instance, \citeauthor{hou2023large} (\citeyear{hou2023large}) represent a user's movie viewing history with a template such as: \textit{``I’ve watched the following movies in the past in order: 1. Multiplicity, 2. Jurassic Park, ... ''}. Alternatively, LLMs can be prompted to summarize such interaction histories \citep{yin2023heterogeneous, zhou2024language, wei2024llmrec}. In either case, such verbalized histories offer an alternative to the sometimes arbitrary mapping of distinct interactions types to numerical or categorical formats, and can also cover pointwise, sequential, and bundle interactions. 




\subsubsection{NL Interactions}
Other user-system interactions are inherently text-based -- for instance, reviews were already discussed in Section \ref{sec:item text}.

\paragraph{Queries} While RSs have traditionally focused on query-less personalization, \textit{queries} -- short text spans expressing a user's real-time information need -- are becoming a key part of language-driven RSs \citep{reddy2022shopping,he2022query}. In fact, this integration of NL queries into RSs is contributing to the convergence of the recommendation and information retrieval (IR) fields.

\paragraph{User-System Dialogue} As conversational recommendation systems (CRS) develop, user-system \textit{dialogues} are emerging as a primary source of textual interaction data \citep{li2018towards}.
Each user and system utterance can reflect diverse intents, including conveying preferences, recommendations, and explanations, as well as asking and answering questions about items or preferences \citep{lyu2021workflow}, as discussed further in Section \ref{sec:ConvRec}.

\subsection{NL User Profiles} \label{sec:NL profiles}
The left side of Figure \ref{fig:ch4_data} illustrates various textual representations of user preferences, including metadata and NL user profiles. 
Recently, \citeauthor{radlinski2022natural} (\citeyear{radlinski2022natural}) proposed that language-driven recommendation could be centered on \textit{scrutable} NL user profiles: textual descriptions of user preferences that are editable and understandable by humans. Such user profiles could succinctly summarize user interests through both specific examples of preferred items 
and generic preference descriptions.

While research on effectively generating, editing, and leveraging these interpretable NL preference representations is still nascent, several initial studies have emerged. For instance, \citeauthor{sanner2023large} (\citeyear{sanner2023large}) have users directly express generic NL item preferences in personal NL profiles, while other authors \citep{yin2023heterogeneous,zhou2024language} use and LLM to generate NL profiles based on item rating histories -- both works find that recommendation performance is competitive with or better than conventional baselines in cold-start settings.   


\paragraph{User Agency via Editable NL Profiles} In principle, an editable NL profile has the potential to empower users to correct system errors, safeguard their privacy, and exert natural control over their preference representations \citep{radlinski2022natural}. It may also be well-suited to handle \textit{preference shifts} \citep{hosseinzadeh2015adapting,pereira2018analyzing} by allowing users to delete obsolete interests or describe temporary contexts. In addition, users can express \textit{aspirations} - desires that do not necessarily align with past behavior, but should influence future recommendations \citep{ekstrand2016behaviorism}. Finally, user edits that result in improved recommendations can incentivize further feedback and increase user satisfaction \citep{bostandjiev2012tasteweights, harper2015putting, knijnenburg2012inspectability}.

\section{Encoder-only LLM Recommendation} \label{sec:encoder-only LLM}

\begin{figure}
    \centering
    \includegraphics[width=\linewidth]{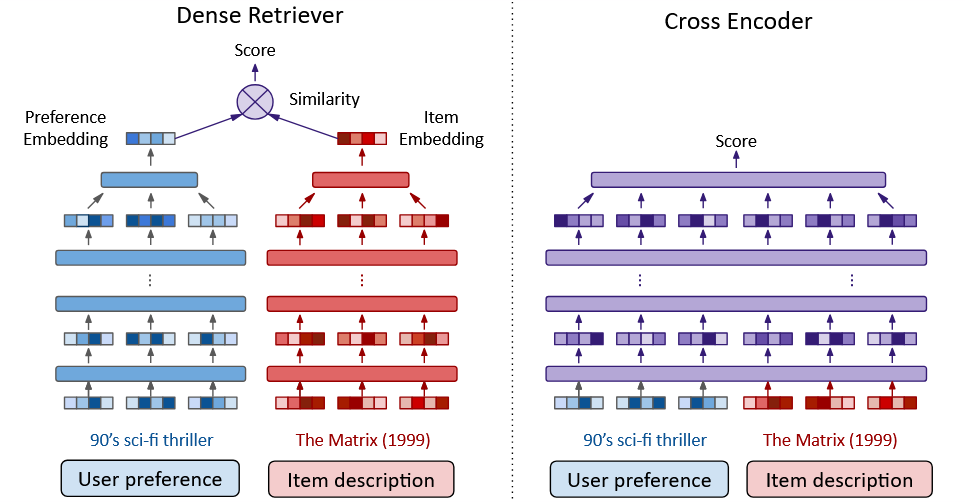}
    \caption{Two main architectures for encoder-only LLM recommendation: a) dense retrievers (left) which encode item and preference descriptions separately and then compute a preference-item embedding similarity score and b) cross-encoders (right) which jointly encode user preference and item descriptions to predict a score.}
    \label{fig:ch4_dense_vs_cross} 
\end{figure}

We now begin our discussion of how textual data can be used in LLM-driven RSs -- starting with recommendation with encoder-only LLMs. As shown in Figure \ref{fig:ch4_dense_vs_cross}, encoder-only LLMs can be used in two main architectures: as dense retrievers (c.f. Sec \ref{sec:dense retrieval}) or as cross-encoders (c.f. Sec \ref{sec:Item-Preference Fusion}). The key difference is that dense retrievers encode preference and item descriptions \textit{separately} while cross-encoders encode them \textit{jointly}, which generally makes cross-encoders slower but more accurate.

\subsection{Dense Retrievers} \label{sec:dense retrieval}
First introduced in the field of information retrieval (IR), dense retrievers
produce a ranked list of documents given a query by evaluating the similarity (e.g., dot product or cosine similarity) between a document embedding and query embedding \citep{fan2022pre}. Dense retrieval is highly scalable (especially with approximate search libraries like
FAISS\footnote{\href{https://github.com/facebookresearch/faiss}{https://github.com/facebookresearch/faiss}}) since document embeddings can be pre-computed and only the query embedding needs to calculated at query time.

To use dense retrieval for recommendation, first, a component of each item's text content, such as its title, description, reviews, etc., is treated as a document. Then, a query is formed by some NL user preference description -- for instance: an actual search query, the user's recently liked item titles, or text generated based on a user utterance in a dialogue \citep{penha2020does}. 

Several recent works explore recommendation as standard dense retrieval, including with off-the-shelf \citep{penha2020does, harte2023leveraging, zhang2023recipe} and fine-tuned \citep{mysore2023large,li2023gpt4rec,hou2023large} retrievers. Dense retrieval is especially common in news recommendation since news articles are very rich in text -- here, a common approach is to aggregate embeddings of a user's recently liked articles into a query embedding, either via mean pooling \citep{iana2023simplifying,iana2024news} or a multi-level encoder (e.g., \cite{wu2019neural, wu2019npa, li2022miner}), before scoring with news candidate embeddings. Another line of work focuses on retrieval using item reviews, and includes studies of contrastive retriever tuning \citep{abdollah2023self} and multi-aspect retrieval \citep{korikov2024multi}. 



\subsection{Cross-Encoders} \label{sec:Item-Preference Fusion}
In contrast to dense retrievers, cross-encoders embed a query and document \textit{jointly}, allowing cross-attention between query and document tokens \citep{fan2022pre}. Several works approach rating prediction by jointly embedding NL item and preference descriptions in LLM cross-encoder architectures with a rating prediction head \citep{zhang2021unbert,yao2022reprbert, wu2021empowering,qiu2021u, zhang2023prompt}. Such fusion-in-encoder methods often exhibit strong performance because they allow interaction between user and item representations, but are much more computationally expensive than dense retrieval
and thus may be best used for small item sets or as rerankers. 

\section{Generative Recommendation and Explanation} \label{sec:LLM gen rec}

\begin{figure}[t]
    \centering
    \includegraphics[width = \textwidth]{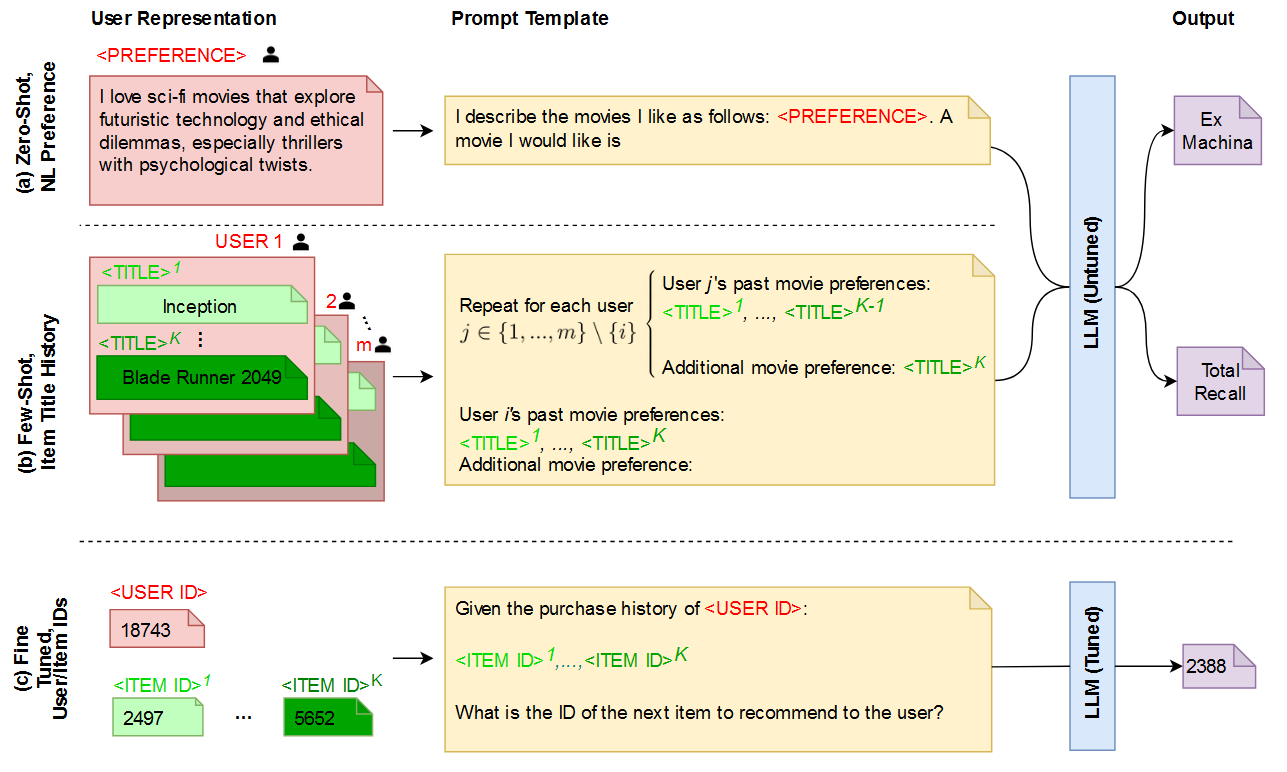}
    \caption{Three examples of generative recommendation with autoregressive LLMs, showing alternative user preference representations and prompting approaches. a) A ZS prompt with a user-generated NL preference description. b) A FS prompt based on sequences of liked item titles from multiple users. c) An ID-based prompt to a tuned LLM which has learned to use item and user ID tokens.} \label{fig:tuned and untuned prompts}
\end{figure}

We next move beyond item-preference scoring to generative recommendation (c.f. Sec \ref{subsec:ICL}-\ref{sec:tuned rec gen}) and explanation (c.f. Sec \ref{sec:gen expl}) with autoregressive LLMs.  Autoregressive LLM inputs are called \textit{prompts}, which are sequences of tokens expressing a task such as top-k recommendation, rating prediction, or explanation generation \citep{geng2022recommendation}. As illustrated in Figure \ref{fig:tuned and untuned prompts}, prompts typically consist of word tokens (Figure \ref{fig:tuned and untuned prompts} a-b), but may also feature non-word tokens such as user and item IDs (Figure \ref{fig:tuned and untuned prompts} c) that can be learned through tuning (c.f. Sec. \ref{sec:tuned rec gen}). While the space of possible LLM inputs and outputs is extremely large, common tasks include generating:
\begin{itemize}
    \item a recommended list of item titles or ids (e.g., \cite{mao2023unitrec, harte2023leveraging, sanner2023large, sileo2022zero}) 
    \item item ratings (e.g. \cite{bao2023tallrec,kang2023llms, zhou2024language})
    \item explanations of recommendations (e.g., \cite{ni2019justifying, li2020generate, hada2021rexplug, geng2022recommendation, li2023personalized}) 
\end{itemize}

This section further discuses such single-turn (i.e., non-conversational), autoregressive LLM recommendation and explanation, covering both untuned  and tuned approaches. 

\subsection{Zero- and Few-Shot Recommendation}\label{subsec:ICL}

The simplest way an autoregressive LLMs can be used is in the \textit{untuned} (i.e., ``off-the-shelf'') setting (e.g. \cite{sileo2022zero,sanner2023large}). As illustrated in Figure \ref{fig:tuned and untuned prompts}, this includes:
\begin{itemize}
    \item zero-shot (ZS) approaches, which rely solely on the LLM's pre-trained knowledge without any additional training data (Figure \ref{fig:tuned and untuned prompts} a)
    \item few-shot (FS) approaches, which provide a small number of input-output examples in the prompt (Figure \ref{fig:tuned and untuned prompts} b), and thus are also referred to as in-context-learning (ICL) methods.
\end{itemize}

\paragraph{Zero- and Few-Shot Prompt Engineering} Clearly, there is a large design space for prompting approaches, including choices for representing user preferences, specifying task instructions, and selecting few-shot examples. Figure \ref{fig:tuned and untuned prompts} illustrates two specific examples from this design space \citep{sanner2023large}: a) a ZS prompt with a user-generated NL preference description, and b) a FS prompt based on sequences of liked movie titles from multiple users. Many other variants are possible, for instance: constructing FS examples based on interactions from the \textit{same} user instead of from different users, including user \textit{dis}preferences, or using LLM-generated user profiles \citep{zhou2024language}. 

\paragraph{Initial Experimental Findings}  Several recent publications have evaluated off-the-shelf LLMs for movie and book recommendation -- domains where relevant knowledge is likely to be internalized during pretraining. Specifically, these methods construct prompts using NL representations of user preferences and instructions to recommend item titles \citep{sanner2023large,sileo2022zero,liu2023chatgpt} or predict ratings \citep{kang2023llms, liu2023chatgpt, zhou2024language}. These initial studies find that while untuned LLMs generally underperform supervised CF methods given sufficient training data \citep{kang2023llms,sileo2022zero}, they are competitive in near cold-start settings \citep{ sileo2022zero, sanner2023large, zhou2024language}. They also suggest that FS typically outperforms ZS prompting and that LLMs struggle with negated reasoning in recommendation (e.g. reasoning about \textit{dis}preferences).

\subsection{LLM Tuning for Generative Recommendation} \label{sec:tuned rec gen} To improve an LLM's generative recommendation performance, multiple works study LLM tuning on historical RS data. First, historical data is converted into textual input-output training pairs for generative recommendation tasks \citep{geng2022recommendation}, which may include:
\begin{itemize}
    \item top-$k$ recommendation (e.g., \cite{geng2022recommendation, li2023prompt}), such as in Figure \ref{fig:tuned and untuned prompts} a-b
    \item sequential recommendation (e.g.,  \cite{harte2023leveraging, li2023prompt, mao2023unitrec}), such as in Figure \ref{fig:tuned and untuned prompts} c
    \item rating prediction (e.g.,\cite{bao2023tallrec,kang2023llms, zhang2023collm})
\end{itemize}
Notably, \citeauthor{geng2022recommendation} \textit{et al.} (\citeyear{geng2022recommendation}) point out that text is a unifying format for training data, enabling multi-task LLM tuning on all of the above tasks. They also show that LLMs can learn to use user and item ID tokens for these tasks (e.g., Figure \ref{fig:tuned and untuned prompts} c). 

Given textual RS training data, the two main approaches to LLM tuning are: 
\begin{itemize}
    \item \textit{fine-tuning}, where training set performance is optimized by adjusting LLM weights \citep{geng2022recommendation, bao2023tallrec, harte2023leveraging, mao2023unitrec, kang2023llms}
    \item \textit{prompt-tuning} where training set performance is optimized by learning prompt tokens (hard prompt-tuning) or embeddings (soft prompt-tuning) \citep{li2023personalized, cui2022m6,zhang2023collm}
\end{itemize}

 
\subsection{Generative Explanation} \label{sec:gen expl} 
Autoregressive LLMs can also generate explanations that aim to help users comprehend why recommendations were made. Similarly to recommendation generation, explanations can be generated by prompting LLMs that are either untuned (e.g., \cite{rahdari2024logic}) or tuned (e.g., \cite{li2023personalized}) -- though the latter is more common for academic publications. To tune LLMs for RS explanation, the typical data source for constructing textual training examples are user reviews, as these are assumed to contain justifications for users' opinions about the reviewed items \citep{ni2019justifying}. 

Broadly, recently studied techniques for RS explanation generation include:
\begin{itemize}
    \item ZS and FS prompting \citep{rahdari2024logic}
    \item fine-tuning \citep{geng2022recommendation, li2023personalized, wang2024deciphering}
    \item prompt-tuning \citep{li2023personalized, li2020generate}
    \item controllable decoding -- where predicted parameters such as ratings steer LLM decoding  \citep{ni2018personalized,ni2019justifying,hada2021rexplug,xie2023factual} 
\end{itemize}

\section{Retrieval Augmented Recommendation} \label{sec:RARec}
\begin{figure}[t]
    \centering
    \includegraphics[width=\linewidth]{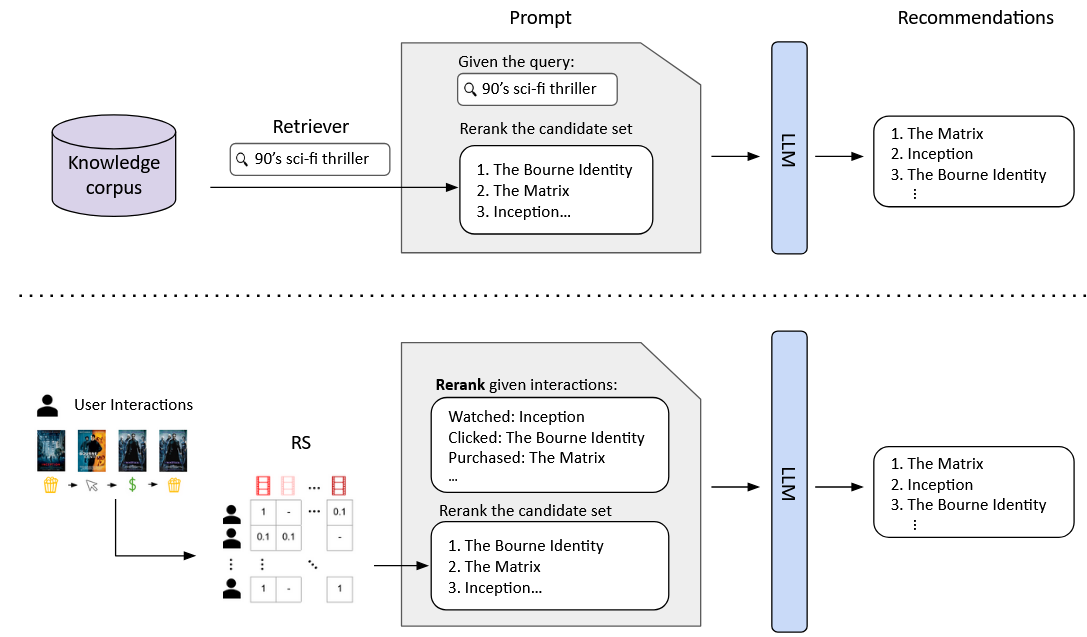}
    \caption{Two examples of RAG for top-$k$ recommendation, where an external tool produces a candidate item set which an LLM is prompted to rerank given a textual description of user preferences. Top: A query is used by a retriever to search for a candidate item set, which an LLM is prompted to rerank given the query. Bottom: A user's interaction history is used by an RS to select a candidate item set, which an LLM is prompted to rerank given the interaction history.} 
    \label{fig:ch4_rag}
\end{figure}

While the previous two sections explore the use of LLM knowledge internalized through pretraining or tuning to generate recommendations and explanations, relying solely on internal LLM knowledge has several limitations \citep{lewis2020retrieval}:
\begin{itemize}
    \item a relatively large number of LLM weights are needed to store knowledge
    \item retraining is needed for each knowledge update
    \item there is no inherent source attribution mechanism
\end{itemize}

To address these limitations, recent work (e.g., \cite{lewis2020retrieval, izacard2020leveraging, borgeaud2022improving, mialon2023augmented}) explores a framework called retrieval-augmented generation (RAG) in which: 1) relevant content is retrieved from an external knowledge source and 2) the retrieved content is used to prompt an autoregressive LLM to generate textual output. These studies provide evidence that, in contrast to approaches relying solely on internal LLM memory, RAG can: reduce the number of LLM parameters (since knowledge can be externalized); provide a convenient mechanism for knowledge updates; improve factuality through better source attribution and by reducing hallucinations.

\subsection{RAG in RSs}
There are many opportunities to use RAG in RSs, including to generate recommendations, explanations, and question answers. More broadly, the RAG framework introduces us to modular architectures for LLM-driven RSs -- a concept we'll explore further when discussing LLM representation generation (c.f. Sec. \ref{sec:LLM 4 RS Inputs}) and conversational RS architectures (c.f. Sec. \ref{sec:ConvRec}).

\paragraph{RAG for Recommendation} As illustrated in Figure \ref{fig:ch4_rag}, the most commonly studied RAG recommendation method has been LLM candidate item set reranking. Specifically, this method involves: 1) selecting a candidate item set based on RS interaction data and 2) prompting an LLM to rerank the candidate set given some information about user preferences \citep{yang2022improving, hou2023large, chen2023palr, wang2023zero, dai2023uncovering, wei2024llmrec}. Recall from Section \ref{sec: NL in RS} that user preferences can be represented in diverse forms, leading to many alternatives for candidate selection and reranking methods. For instance, Figure \ref{fig:ch4_rag} a) illustrates candidate selection with a retriever driven by a user query, and LLM candidate reranking given this query. As another example, Figure \ref{fig:ch4_rag} b) shows candidate selection with an RS based on item interaction history, and LLM reranking given this history.  

\paragraph{RAG for Explanation and Conversational Recommendation} Other examples of how RAG can be used in RSs include explanation generation and as a tool for conversational recommendation. For RAG-based explanation generation, \citeauthor{xie2023factual} (\citeyear{xie2023factual}) generate queries based on interaction history to retrieve an item's reviews, which are then used as context to generate an explanation of the recommendation. Examples of RAG in conversational recommendation, discussed further in Section \ref{sec:ConvRec}, include work by \citeauthor{friedman2023leveraging} (\citeyear{friedman2023leveraging}) to retrieve relevant user preference descriptions from a user ``memory'' module, and by \citeauthor{kemper2024retrieval} (\citeyear{kemper2024retrieval}) to retrieve information from an items reviews to answer user questions. 

\section{LLMs Representation Generation} \label{sec: LLM 4 RS Inputs}
\begin{figure}
    \centering
    \includegraphics[width=0.75\linewidth]{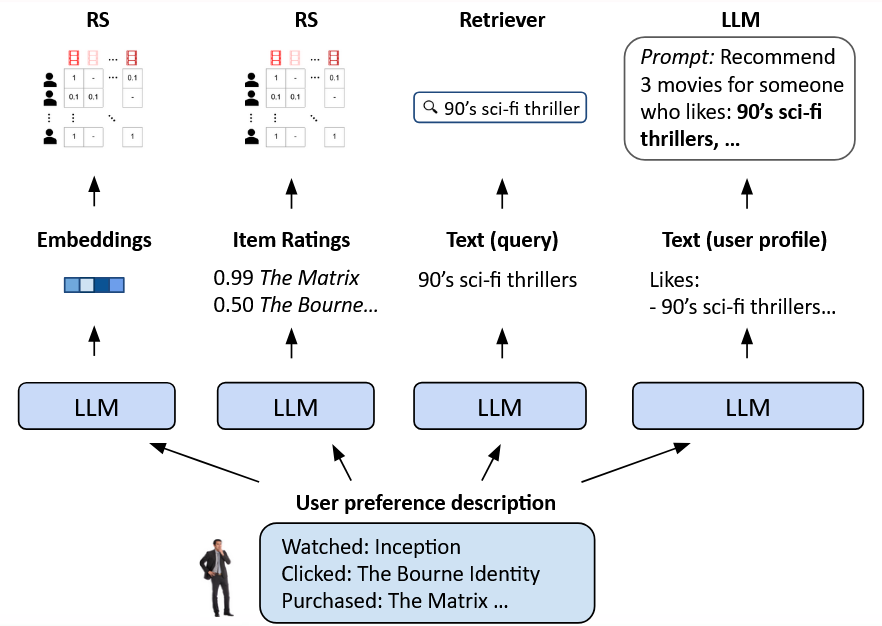}
    \caption{Examples of LLM generated-inputs for downstream modules, including embeddings and predicted item ratings for downstream RSs, and text for downstream search or LLM prompting.}
    \label{fig:ch4_llm_feature_gen.png}
\end{figure}


While Section \ref{sec:RARec} explored using an external module such as retrievers or RSs to select LLM inputs, this section discusses the converse setting in which LLMs generate inputs to downstream modules, as illustrated in \ref{fig:ch4_llm_feature_gen.png}. Specifically, LLMs can be used to transform text such as item and preference descriptions into representations (e.g., embeddings, text, item ratings) that serve as inputs to modules such as RSs, retrievers, or autoregressive LLMs. Such text transformations can be interpreted as an LLM encoding step \citep{zhou2024language}, including for the case of text-to-text transformations -- such as the generation of user preference profiles based on user history.

\subsection{Text to Text}
Two common settings where an LLM generates text that serves as input to a downstream module are search queries and LLM prompt elements, with several example studies for both approaches discussed below.

\paragraph{Search Queries} Given a user's item interaction history (and item text), MINT \citep{mysore2023large} 
and GPT4Rec \citep{li2023gpt4rec} prompt an LLM to generate an search query for a dense retriever with the goal of finding new items to recommend. Similarly, recent work on conversational recommendation uses dialogue history to generate search queries to find items to recommend \citep{friedman2023leveraging,kemper2024retrieval}. 

\paragraph{Prompt Elements} As mentioned in Section \ref{sec:NL profiles}, recent work \citep{yin2023heterogeneous,zhou2024language} first uses an LLM to generate a NL user profile based on their interaction history, then prompts an LLM to generate recommendations given this profile. In conversational recommendation, LLMs are often employed to generate prompt elements (or entire prompts) for downstream tasks based on dialogue history \cite(e.g., \cite{wang2022unicrs, friedman2023leveraging, gao2023chat, kemper2024retrieval}). 


\subsection{Text to Embeddings}
As discussed in Chapters 2 and 3, many RS techniques rely on latent representations of items and user preferences. Thus, a recent line of research uses LLMs to encode semantic information from text into latent embeddings which are then used for recommendation. While encoder-only LLM recommendation was already discussed in Section \ref{sec:encoder-only LLM}, this section covers multi-stage pipelines where LLM embeddings are used for downstream recommendation modules, discussed next. 

\paragraph{Sequential Recommendation with LLM Embeddings} Recall that Chapter 3 discussed several attention-based sequential recommendation models (e.g., BERT4Rec, SASRec) which predict the next item ID to recommend given a sequence of item IDs that a user has interacted with. Recent work uses LLMs to initialize item ID embeddings based on item text \citep{harte2023leveraging, yuan2023go, rajput2024recommender}, reporting significant performance gains. Similarly, Query-SeqRec \citep{he2022query} incorporates user query information into sequential recommendation by using an LLM to encode queries alongside item embeddings.

\paragraph{Rating Prediction with LLM Embeddings} LLMs have also been used to generate latent representations of items (based on item text) and users (based on item interactions) that serve as inputs to a neural network which predicts a user-item rating \citep{wu2021empowering, yuan2023go}. These approaches essentially augment the well-known neural collaborative filtering \citep{he2017neural} framework with LLM embeddings. 

\subsection{Text to Item Ratings} 
As discussed further in Section \ref{sec:ConvRec}, LLMs have been used to map dialogue history to item rating predictions, with these ratings then treated as observations by an RS to make recommendations. Examples include sentiment analysis towards items mentioned in a dialogue \citep{li2018conversational} and natural language inference between user-stated aspect preferences and item text \citep{austin2024bayesian}.

\section{Conversational Recommendation} \label{sec:ConvRec}
The previous sections mostly focused on single-turn LLM recommendation personalized based on \textit{pre-existing} user history information, such as non-verbal interactions, queries, reviews, and so on (c.f. Figure \ref{fig:ch4_data}), and item text. However, LLMs provide new opportunities for multi-turn conversational recommender systems (CRSs) where each turn presents a chance for the user to clarify or revise their preferences, critique and ask questions about recommended items, or convey a variety of other real-time intents such as those shown in Table \ref{tab:ch4_user_intents} \citep{lyu2021workflow}.  Correspondingly, CRSs should facilitate a wide range of responses
including revising recommendations, responding to questions, and personalizing explanations. Further, each turn is also an opportunity for the CRS to generate proactive utterances such as clarifying questions or explanations focused on key topics to help elicit user preferences. As with user intents, various possible CRS actions are summarized in Table \ref{tab:ch4_sys_intents} \citep{lyu2021workflow}. LLM-driven CRSs thus present opportunities not only to personalize recommendations, but also to personalize system interactions more broadly. 

\begin{table*}[t]
 \caption{The user intent conversational recommendation taxonomy of \citeauthor{lyu2021workflow} (\citeyear{lyu2021workflow}), with $\dag$ indicating a additional category to the taxonomy of \citeauthor{cai2020predicting} (\citeyear{cai2020predicting}).}
  \rowcolors{2}{gray!10}{white}
  \resizebox{\linewidth}{!}{%
  \begin{tabular}{lll}
  \toprule
    \textbf{Category} &  \textbf{Description} & \textbf{Example} \\
  \midrule
    \textbf{Ask for Recommendation} & &  \\
    \hspace{3mm}Initial Query & User asks for a recommendation in the first query & ``Hi I am looking for a place to have a family brunch...'' \\
    \hspace{3mm}Continue & User asks for another recommendation in a subsequent query & ``Maybe you can give me one more choice so I can pick one...'' \\
    \textbf{Provide Preference} & & \\
    \hspace{3mm}Provide Context $\dag$ & User provides background information for the restaurant search & ``I am looking for a restaurant for my Valentine’s day dinner.'' \\
    \hspace{3mm}Provide Preference & User provides specific preference for the desired item & ``I would prefer a place that has a very good scenic view.'' \\
    \hspace{3mm}Refine Preference $\dag$ & User improves over-constrained/under-constrained preferences & ``It does not have to be chicken fingers.'' \\
    \textbf{Answer} & User answers the question issued by the recommender & ``Yes that’s correct.'' \\
    \textbf{Acknowledgement $\dag$} & User shows understanding towards a previous recommender utterance & `` I see.'' \\
    \textbf{Recommendation Rating} & & \\
    \hspace{3mm}Been to (modified) & User has been to the restaurant before & ``Oh I have been there before.'' \\
    \hspace{3mm}Accept & User accepts the recommended item, either explicitly or implicitly & ``Ok our final choice will be Eggspectation.'' \\
    \hspace{3mm}Reject & User rejects the recommended item, either explicitly or implicitly & ``Maybe there is a private room in the other three restaurants?'' \\
    \hspace{3mm}Neutral Response  & User does not indicate a decision with the current recommendations & ``I will take a look in the menu and compare and maybe ask my partner.''\\
    \textbf{Inquire }  & User requires additional information regarding the recommendation & ``So what about the interior design, the decorations and environment?'' \\
    \textbf{Critiquing} & & \\
    \hspace{3mm}Critique - Feature  & User critiques on a specific feature of the recommended item & ``I am pretty sure it will be expensive so what is the price range?''\\
    \hspace{3mm}Critique - Add  & User adds further constraints on top of the current recommendation & ``I want sushi.'' \\
    \hspace{3mm}Critique - Compare  & User requests comparison between recommended item with another item  & ``How about the price compared with Miku??'' \\
    \textbf{Others } & Greetings, gratitude expression, chit-chat utterance & ``Thank you so much for your recommendation.'' \\
       \bottomrule

  \end{tabular}
}%
 \label{tab:ch4_user_intents}
\end{table*}

\begin{table*}[t]
 \caption{The recommender intent conversational recommendation taxonomy of \citeauthor{lyu2021workflow} (\citeyear{lyu2021workflow}), with $\dag$ indicating a additional category to the taxonomy of \citeauthor{cai2020predicting} (\citeyear{cai2020predicting}).}
 \rowcolors{2}{gray!10}{white}
  \resizebox{\linewidth}{!}{%
  \small
  \begin{tabular}{lll}
  \toprule
    \textbf{Category} &  \textbf{Description} & \textbf{Example}  \\
  \midrule
    \textbf{Request} & & \\
    \hspace{3mm}Request Information  & Recommender requests the user’s preference & ``What kind of food do you like?'' \\
    \hspace{3mm}Clarify Question  & Recommender asks for clarification on a previous requirement & ``So you would like to reserve a private room?''\\
    \hspace{3mm}Ask Opinion $\dag$ & Recommender requests the user’s opinion to a choice question (e.g., yes/no) &  ``So it is just open space but separated from others, is that ok?'' \\
    \hspace{3mm}Ensure fulfillment $\dag$ & Recommender confirms task fulfillment during the conversation & ``Anything else I can do for you today?'' \\
    \textbf{Inform progress $\dag$} & Recommender discloses the current item being processed & ``So let me just check the closest nearby parking.''\\
    \textbf{Acknowledgement $\dag$} & Recommender shows understanding towards a previous user utterance & ``...you mentioned that one of the attendees is vegetarian...''\\
    \textbf{Answer } & Recommender answers the question issued by the user & ``So for the Michael’s on Simcoe, the price varies a lot...'' \\
    \textbf{Recommend} & & \\
    \hspace{3mm}Recommend - Show  & Recommender provides recommendation by showing it directly & ``So I found a restaurant called paramount.'' \\
    \hspace{3mm}Recommend - Explore  & Recommender provides recommendation and asks if the user has prior knowledge & ``The first one that comes to mind is Miku, have you heard of it before?'' \\
    \textbf{Explain} & & \\
    \hspace{3mm}Preference  & Recommender explains recommendations based on the user’s said preference & ``Because it has vegetarian options, it has a full bar and a good view...'' \\
    \hspace{3mm}Additional Information $\dag$ & Recommender explains recommendations with features not previously discussed & ``There are a couple different varieties (of food) that your guests might enjoy.'' \\
    \hspace{3mm}\textbf{Personal Opinion} & & \\
    \hspace{6mm}Comparison $\dag$ & Recommender compares recommended item with another item &  ``I would say the price for this place is a bit higher than HY steakhouse but...'' \\
    \hspace{6mm}Persuasion $\dag$ & Recommender provides positive comment towards the recommended item & ``It is on the pricier side but it is worth the experience...'' \\
    \hspace{6mm}Prior Experience $\dag$ & Recommender refers to past experience with the recommended item & ``I have been there before during the summerlicious.'' \\
    \hspace{6mm}Context $\dag$ & Recommender provides opinion considering the given context or current reality & ``Since it’s summer I don’t think the weather will be that much of an issue...'' \\
    \textbf{Others } & Greetings, gratitude expression, chit-chat utterance & ``Yeah a lot of people recommended me to go there.'' \\
   \bottomrule
  \end{tabular}
  }%
 \label{tab:ch4_sys_intents}
\end{table*}

\paragraph{Pre-LLM CRS Architectures} Historically, most pre-LLM CRSs followed a two-step process in each turn: 1) \textit{belief tracking} to maintain a dialogue state, and 2) \textit{response generation} to produce a system utterance \citep{jannach2020survey}. 
Belief tracking was often implemented with slot-filling techniques, where the dialogue history would be used to update a \textit{dialogue state} consisting of variables, or slots (e.g., ``\textit{preferred\_cuisine: \_}''), that were filled from a set of predefined values (e.g., ``\textit{Mexican}'', ``\textit{French}'', ...) \citep{williams2014dialog,budzianowski2018multiwoz}. These slot-based states would then be used to determine the system response, which often used NL templates and item metadata. However, these slot-based architectures exhibited limited NL reasoning capabilities, constraining their capacity to understand and represent complex user intents and generate deeply personalized responses and recommendations. 

\begin{figure}[t]
   \centering
   \includegraphics[width=\linewidth]{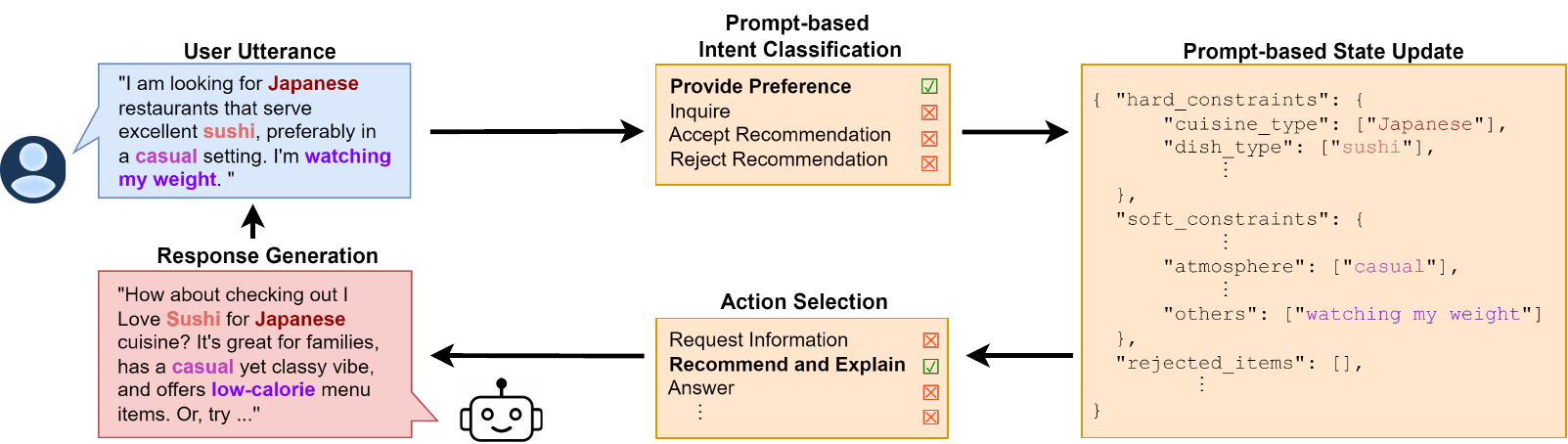}
   \caption{The \textit{RA-Rec} prompt-driven dialogue management approach \citep{kemper2024retrieval}. LLM prompting is used to maintain a JSON semi-structured NL state which tracks user preferences and intents. The predefined state keys (e.g., \textit{``cuisine\_type''}) provide a configurable structure, while the LLM-generated state values can dynamically express nuanced concepts in NL (e.g., ``\textit{``watching my weight''}).}
   \label{fig:ch4_JSON}
\end{figure}

\paragraph{LLM-Driven CRSs}
In contrast to highly constrained slot-based dialogue states described above, LLMs enable rich textual representations of the dialogue state. For instance, if we consider the basic case of using a monolithic LLM \textit{as} a CRS (e.g., \cite{he2023large}), the dialogue state simply becomes the conversation history, which preserves the full dialogue context. As discussed further in Section \ref{sec:ch4_beliefs}, recent CRS research often explores more complex text-augmented dialogue states -- expanding or replacing the raw conversation history with elements such as JSON-like dialogue states (e.g., Figure \ref{fig:ch4_JSON}), user preference summaries \citep{friedman2023leveraging,joko2024doing}, or numerical variables (e.g., inferred item ratings). These text-augmented dialogue states are then used to form inputs to response generation modules, which may include not only LLMs but also tools such as retrievers, RSs, and KGs, as discussed in Section \ref{sec:system-response-generation}.   


\subsection{Belief Tracking} \label{sec:ch4_beliefs}
LLMs enable a CRS to track its beliefs about the conversation through text-augmented dialogue states. This section covers the use of purely textual components in dialogue states as well as the incorporation of non-textual elements alongside text. Both textual and non-textual dialogue state elements can form inputs to response generation modules, discussed further in Section \ref{sec:system-response-generation}.

\subsubsection{Textual Dialogue State Components}
The most basic form of a textual dialogue state is simply the conversation history -- the default state representation when a monolithic LLM is used as a CRS (e.g., \cite{he2023large}). However, the pure conversation history can also be extended or substituted with other textual state components. For instance, \citeauthor{kemper2024retrieval} (\citeyear{kemper2024retrieval}) prompt an LLM to convert the conversation history into a JSON dialogue state, as shown in Figure \ref{fig:ch4_JSON}, which provides a semi-structured format to represent beliefs about the user's item preferences (e.g., preferred cuisine type) and the dialogue flow (e.g., whether the user is expecting an answer to a question). Other authors \citep{friedman2023leveraging,joko2024doing} prompt an LLM to generate textual user preference \textit{memories,} and store these memories as documents in a long-term memory corpus.

\subsubsection{Semi-textual Dialogue States}
In addition to textual components such as those described above, other works also feature categorical or numerical variables in the dialogue state. Specifically, the conversation history can be used to assign values to variables such as:
\begin{itemize}
    \item liked/disliked item titles \citep{liu2020towards} or item categories \citep{huang2023recommender}
    \item mentioned item-related KG entities \citep{chen2019towards,zhou2020towards, liu2020towards, ma2020cr, wang2022unicrs}
    \item user constraints in a constraint reasoning-based CRSs \citep{zeng2024automated}
\end{itemize}
Dialogue states can also include numerical variables -- for instance \citeauthor{li2018towards} (\citeyear{li2018towards}) use a sentiment classification module to infer a user rating towards mentioned items, while \citeauthor{austin2024bayesian} (\citeyear{austin2024bayesian}) maintain Bayesian beliefs over user-item preferences.


\subsection{System Response Generation} \label{sec:system-response-generation}
The primary purpose of the dialogue state elements discussed above is to guide the system response by forming inputs to response generation modules. For instance, textual state components may be used for LLM prompts and/or retriever queries, while non-textual state elements such as item ratings may function as arguments to tools such as RSs. Broadly, LLM-driven CRS response generation can be guided through prompting, tuning, and interfacing with external tools. 

\subsubsection{Prompting}
There are many diverse ways that prompting can be used to guide CRS utterance generation. Several works study \textit{single-intent} systems, where the prompt is constructed using the dialogue history and an instruction to generate an utterance with a fixed intent, such as to recommend \citep{he2023large} or ask a preference elicitation query \citep{handa2024bayesian,austin2024bayesian}. Other methods \citep{joko2024doing, kemper2024retrieval} express hand-crafted dialogue rules through prompts, asking an LLM to select the best system action given the dialogue state and some rules (e.g., ``\textit{Ask a user for their location if they have not provided it, otherwise, recommend a restaurant.}''). In addition, prompts can be partially or fully system generated, for instance using the output of another LLM module, a retriever, or a reasoning tool such as an RS or KG (e.g., \cite{zhou2020towards, wang2022unicrs, friedman2023leveraging, gao2023chat}).  

\subsubsection{Tuning}
CRS dialogue policies can also be controlled by tuning LLMs on human-human or synthesized conversation data \citep{li2018towards, kang2019recommendation, zhou2020towards, liu2020towards, ma2020cr, li2022user, wang2022unicrs, friedman2023leveraging,joko2024doing}. While these methods cover many tuning approaches, they all lead to an LLM internalizing some knowledge about how to respond to the user in NL or execute some reasoning step such as a search or recommendation. 

\subsubsection{Tool Use}
LLM-driven CRS response generation can also be augmented by the use of external tools such as retrievers or RSs. The techniques for interfacing LLMs with such tools -- discussed in Sections \ref{sec:RARec}-\ref{sec: LLM 4 RS Inputs} -- remain applicable in multi-turn settings, with the dialogue now state serving as the basis for representing tool inputs and outputs. In the context of retrieval-augmented CRSs, recent work \citep{friedman2023leveraging, kemper2024retrieval} generates a search query based on preferences represented in the dialogue state to retrieve relevant items based on item text. Similarly, RS modules are used to make recommendation based on inferred item ratings \citep{li2018towards,austin2024bayesian} and recognized KG entities (e.g., \cite{chen2019towards,wang2022unicrs}).

\backmatter

\bibliographystyle{nowfntsort}
\bibliography{sample-now}


\end{document}